\documentclass[10pt,twocolumn,letterpaper]{article}

\usepackage{cvpr}
\usepackage{times}
\usepackage{epsfig}
\usepackage{graphicx}
\usepackage{amsmath}
\usepackage{amssymb}
\usepackage{float}
\usepackage{booktabs} 
\usepackage{multirow}
\usepackage{blindtext}
\usepackage{caption}
\usepackage{authblk}

\usepackage[breaklinks=true,bookmarks=false]{hyperref}

\cvprfinalcopy 

\def\cvprPaperID{2983} 

\begin{document}

\title{Deep Image Category Discovery using a Transferred Similarity Function}
\author[1]{Yen-Chang Hsu}
\author[1]{Zhaoyang Lv}
\author[2]{Zsolt Kira}
\affil[1]{Georgia Institute of Technology}
\affil[2]{Georgia Tech Research Institute}
\affil[ ]{\tt\small {\{yenchang.hsu, zhaoyang.lv, zkira\}@gatech.edu}}

\newcommand{\ZL}[1]{\textcolor{blue}{ZL: #1}}
\newcommand{\YH}[1]{\textcolor{red}{YH: #1}}
\newcommand{\ZK}[1]{\textcolor{green}{ZK: #1}}

\newcommand{\image}[1]{I_{#1}}
\newcommand{\images}{I_{\{1...N\}}}
\newcommand{\cluster}[1]{\theta_{#1}}
\newcommand{\clusters}{\Theta_{\{1...M\}}}
\newcommand{\weights}{\mathbf{w}}
\newcommand{\distribution}{p}
\newcommand{\out}[1]{z_{#1}}
\newcommand{\neuralnetwork}[1]{f({#1})}
\newcommand{\loss}[1]{\mathcal{L}({#1})}
\newcommand{\hingeloss}[1]{L_{h}(#1)}
\newcommand{\KL}[1]{\mathcal{D}_{KL}{(#1)}}
\newcommand{\bigO}[1]{\mathcal{O}({#1})}


\twocolumn[{%
	\renewcommand\twocolumn[1][]{#1}%
	\maketitle
	\begin{center}
		\centering
		\includegraphics[trim={0.3cm 0.5cm 2cm 0.5cm},clip,width=0.77\textwidth]{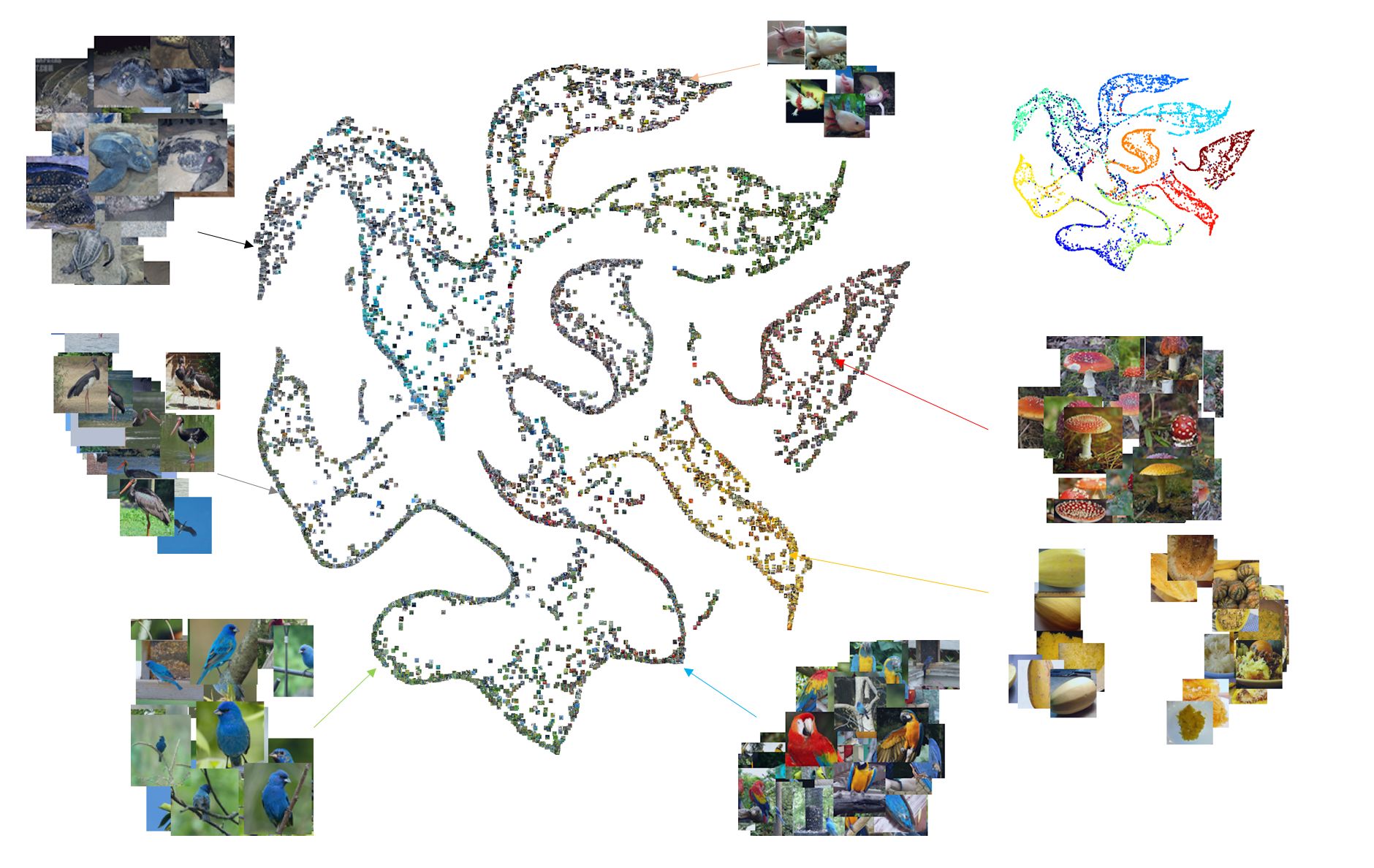}
		\captionof{figure}{Visualization of unuspervised discovery of object categories in natural images via a transferred similarity function (visualized using t-SNE based on our outputs). The input images are from ten categories without labels: Stingray, Indigo bunting, Axolotl, Tree frog, Turtle, Macaw, Black stork, Weimaraner Agaric, and Sphagatti squash.  We also show the embedding with ground truth labels on the top right. An in-depth discussion and quantitative results is included in section \ref{sec:experiments}. }
		\label{fig:imagenet_visualization}
\end{center}%
}]

\begin{abstract}
Automatically discovering image categories in unlabeled natural images is one of the important goals of unsupervised learning. However, the task is challenging and even human beings define visual categories based on a large amount of prior knowledge. In this paper, we similarly utilize prior knowledge to facilitate the discovery of image categories. We present a novel end-to-end network to map unlabeled images to categories as a clustering network. We propose that this network can be learned with contrastive loss which is only based on weak binary pair-wise constraints. Such binary constraints can be learned from datasets in other domains as transferred similarity functions, which mimic a simple knowledge transfer. We first evaluate our experiments on the MNIST dataset as a proof of concept, based on predicted similarities trained on Omniglot, showing a 99\% accuracy which significantly outperforms clustering based approaches. Then we evaluate the discovery performance on Cifar-10, STL-10, and ImageNet, which achieves both state-of-the-art accuracy and shows it can be scalable to various large natural images.
\end{abstract}
\section{Introduction}

Given a set of natural images without labels, humans have the ability to find high-level structure and discover new concepts inside of them. However, it is still a challenging task for machines to mimic in a completely unsupervised manner because humans use the transfer of prior knowledge from other domains. This ability in human learning motivates us to think about the following problem: how can we discover new object categories and structures in unlabeled data by weakly transferring knowledge from other domains?

One form of weak prior knowledge is a similarity function that can be used to make comparisons between images, from which we can generalize to create the set of categories that form the space. Such a process can be traditionally addressed in two stages: predict similarity through metric learning and generate categories based on these metrics by the application of clustering algorithms. However, we argue that such two-process pipelines are both unnatural and biased. Metric learning can induce strong human bias via assumptions which are not necessarily appropriate for this task. Clustering based on such metrics does not naturally utilize the features which exist inherently in unlabeled data.

We posit that end-to-end learning from unlabeled natural images is preferable, using very simple dense binary constraints as input. The binary constraint is simply whether a pair of images is \emph{similar} or \emph{dissimilar}, which can be predicted through knowledge transfer. In other words, we transfer a learned similarity metric to the target domain, but use it to jointly learn feature embedding and a clustering by applying the metric on unlabeled data.  Although such predictions can be inconsistent because of domain differences, we are able to correct such inconsistencies by robustly extracting high level categories via clustering.

In this paper, we propose an end-to-end deep model following this intuition.  We focus on the problem of discovering image categories from the unlabeled images. We address the object category discovery task as a two-step training task: first we train the similarity prediction network (SPN) on an existing dataset as a Siamese architecture to predict the similarity label between two images based on their category labels; then on the new unlabeled data, we train the category discovery as a purely data-driven, end-to-end unsupervised clustering problem whose outputs are expected to be distinguishable distributions. Note that the SPN is directly used as transferred knowledge, and is not updated through learning in the new unsupervised task. The clustering network, however, jointly learns features as well as a clustering output for the new task using unlabeled data and outputs from the fixed SPN.


In transferring a similarity function across domains, there is a potential concern: How will the discovery task perform when the pair-wise predictions are potentially noisy?  We posit that this concern can be greatly alleviated by using {\it dense} pair-wise constraints. Given such dense constraints between images, the network is able to discover structure even though the constraints predicted through knowledge transfer are very noisy. We evaluate this idea with in-depth experiments, and present significant improvement in various datasets, compared to both traditional and state-of-art unsupervised methods.


In our work, we have made the following contributions: 1) We propose an end-to-end deep clustering algorithm to achieve unsupervised image category discovery, with weakly predicted pair-wise constraints learned on datasets from a different domain, 2) Backed by our experiments, we show that the object-level categories in unlabeled images can then be discovered with high accuracy, given noisily predicted dense pair-wise constraints between raw images, and 3) Our results show that this framework is able to achieve the state-of-art clustering performance, and is highly scalable on large scale datasets.


\section{Related Work}

\begin{figure}[t] 
\includegraphics[width=0.45\textwidth]{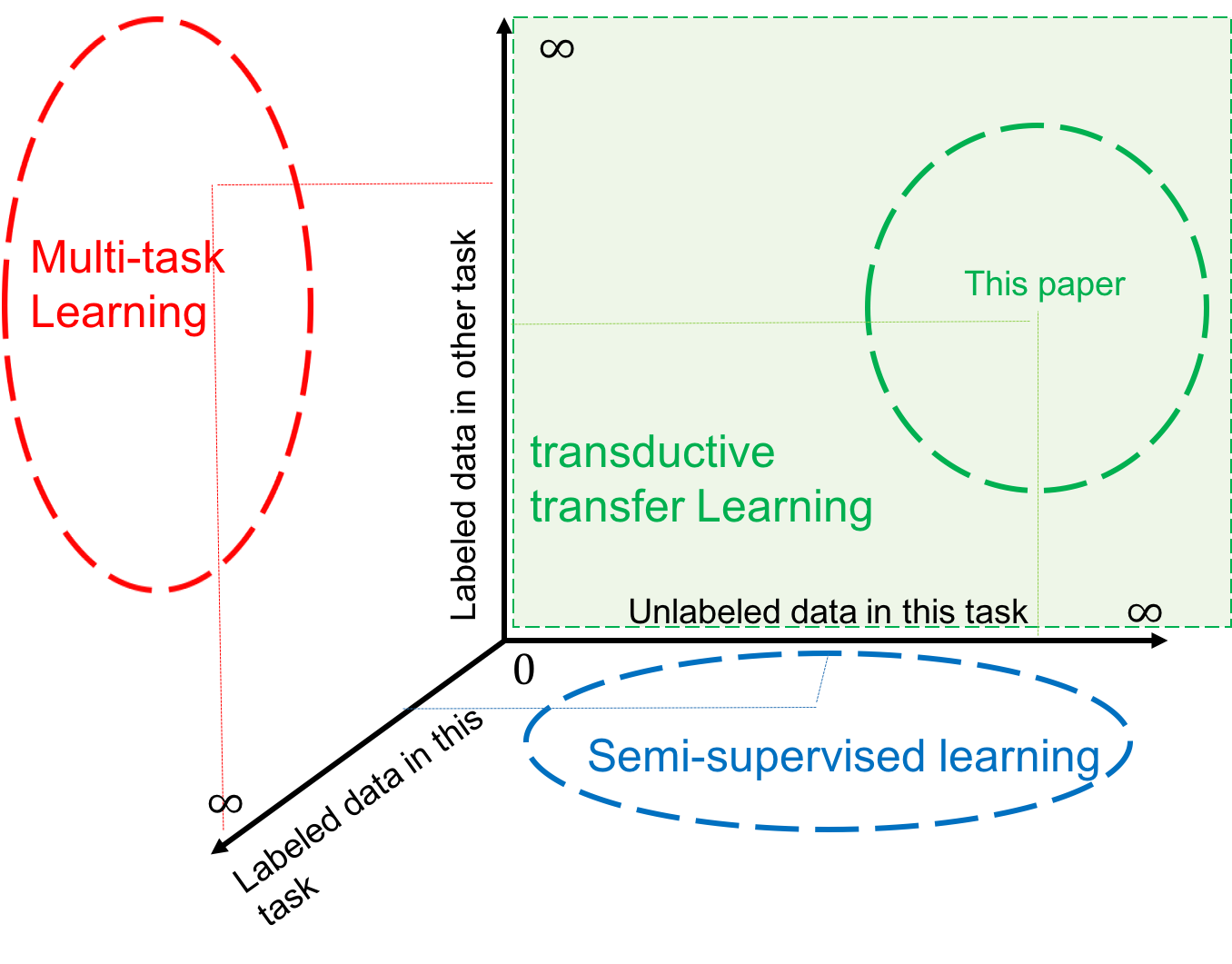}
\caption{The domain of this paper compared to the three major comparable areas in the literature: unsupervised learning, transfer learning and semi-supervised learning. In our work, we propose to use the labeled data from a source task to weakly transfer a learner, which can then be used directly in the new domain task unsupervised, which is termed as transductive transfer learning \cite{Pan10pami} }
\label{fig:problem_domain}
\end{figure}


In this paper,  we address the problem of unsupervised learning with weak transfer of knowledge. The domain of our problem is highlighted as figure \ref{fig:problem_domain}. A variety of work is relevant and inspired our thinking, although not sharing the same problem setting. We summarize them as follows: 

\subsubsection*{Transfer Learning}

In the survey paper of transfer learning \cite{Pan10pami} our method can be termed as transductive transfer learning, in which the source domain labels are available, but target domain labels are unknown. Most of the approaches in this domain address the problem when feature spaces from source task to domain task is similar, which is often addressed in deep learning by sharing feature representation from neural network \cite{Donahue14icml}\cite{Oquab14cvpr}\cite{Razavian14cvprw}\cite{Yosinski14nips}.

In our problem, we propose that by only transferring weak knowledge is enough to achieve equal or even better unsupervised learning performance in the target domain, without assuming source domain and target domain share similar feature representations. Our similarity prediction network (SPN) is used with fixed parameters in the target domain. In \cite{Li16eccv}, Li et al. discussed how new tasks can benefit existing tasks through transfer learning. Different from ours, their focus was on multi-task learning in a supervised domain. Our SPN used in category similarity prediction is similar to \cite{Han15cvpr} and \cite{Zagoruyko15cvpr}, which predict patch similarity of image pairs. Such networks trained via supervised learning demonstrate that it can be generally applied across tasks and scenarios. 


\subsubsection*{Unsupervised Learning as Clustering} 
Clustering-based unsupervised learning approaches can be traditionally posed as a two-stage solution: learning a feature embedding \cite{Donahue14icml} and clustering based on certain metrics \cite{Macqueen67bsmsp}\cite{Shi97pami}\cite{Jain99acs}\cite{Cai09ijcai}. Both are important parts for unsupervised clustering on natural images, but these two steps are typically not tackled in a unified way.There are only a few recent approaches that have explored end-to-end approaches \cite{Yang16cvpr}\cite{Xie16icml}. \cite{Yang16cvpr} proposed joint learning of feature representations and showed that agglomerative clustering can be viewed as a recurrent framework. In \cite{Xie16icml} the network learns a mapping from the data space to a lower-dimensional feature space and iteratively optimizes a clustering objective. However, these approaches are not able to deal with even simple natural images, as evaluated in our experiments. Further, they are not scalable to large-scale image sets as our proposed method is. Recently there are also approaches from the Generative Model perspective and experiments have shown that clustering can be achieved as learning latent embeddings during adversarial learning \cite{Makhzani15arXiv}. 


Our work is also relevant to \cite{Hsu16iclrw}, which is a supervised approach for end-to-end clustering. They proposed an end-to-end clustering framework with a contrastive loss by utilizing supervised ground truth labels. We achieve fully unsupervised clustering by using a learned noisy SPN. We also present our method on natural image discovery, which is hardly tackled in all of the other clustering approaches. 


\subsubsection*{Object Discovery} 

This work is relevant to Object Discovery \cite{Bolanos15arXiv} and \cite{Cho15cvpr} but we do not deal with object localization. A similar problem setting is addressed in \cite{Li16eccv}, which tries to discover unsupervised feature representations by exploiting internal constraints within the unlabeled images. However, this method focuses on feature learning instead of exploring categories present in the data.

%
%
%
\section{Approach}

\begin{figure}[t]
	\centering
	\includegraphics[width=0.8\linewidth]{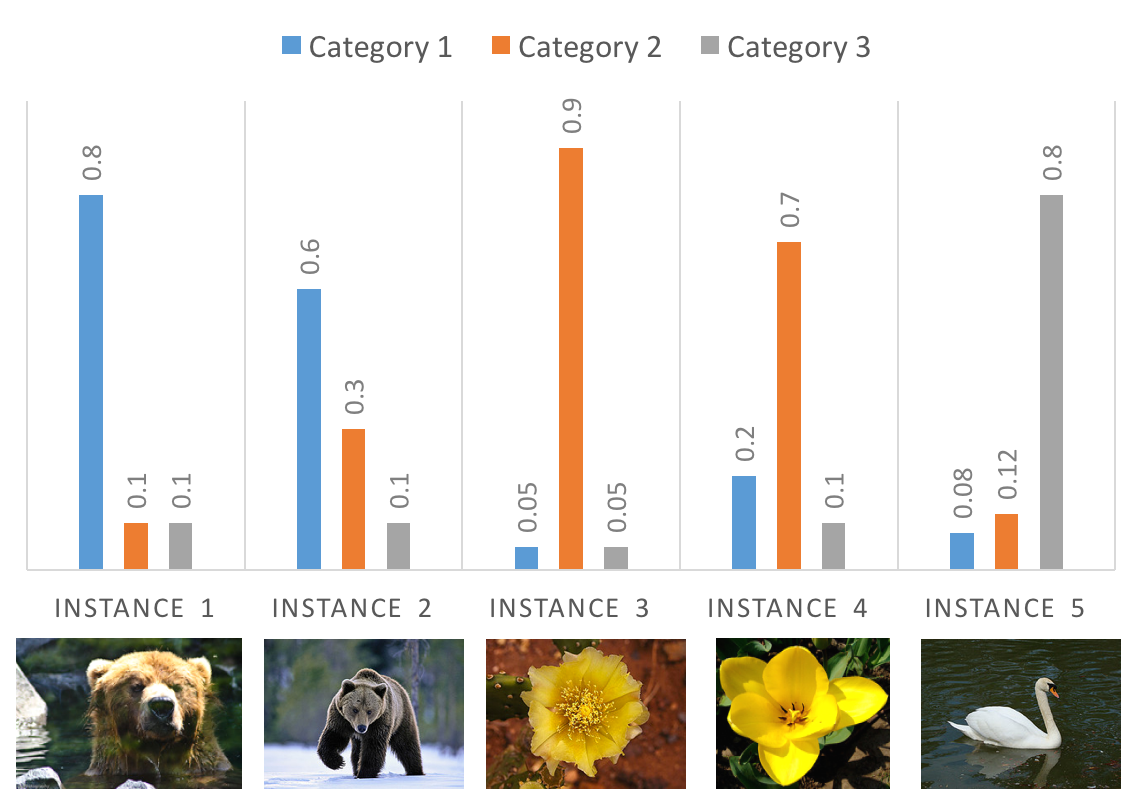}
	\caption{An example depicting our concept of unsupervised category discovery from category distributions. Instances belonging to the same category should have similar output distributions, while instances belonging to different categories should have different distributions. }
	\label{fig:softmax_distribution_example}
\end{figure}

We formulate the problem as follows: Given an unlabeled image set with $N$ images ($\images=\{\image{1}, \image{2}, ..., \image{N} \}$) we expect to discover $M$ image categories $\clusters$ from this image set. For each data instance, we specifically aim to output a distribution over these categories, i.e. $\clusters=\{\cluster{1}, \cluster{2}, ..., \cluster{M} \}$. We approach this by transforming each input image $\image{i}$ into an embedding $\out{i}$. The transformation is learned through a deep convolution neural network, which can be defined as $\out{i} = \neuralnetwork{\image{i}; \weights}$, in which the network weights $\weights$ are learned through optimizing the clustering loss with pairwise similarity. Then a mapping from a cluster to the corresponding category can be obtained through Hungarian algorithm, which is defined in section \ref{sec:eval_metric}.




\subsection{Predicting Cluster Assignment Distributions using Dense Pair-Wise Constraints}

The main intuition of our end-to-end clustering approach is \emph{category mapping}: We expect the transformed output $\out{i}$ from image $\image{i}$, which corresponds to a particular category, to be close to that of all of the images $\image{j}$ corresponding with similar output $\out{j}$. Conversely, images $\image{i}$ and $\image{j}$ from different categories will be ideally mapped to different $\out{i}$ and $\out{j}$ that are far apart.  We show an example of five input images being mapped to three categories in figure \ref{fig:softmax_distribution_example}.

\begin{figure}[t] 
	\centering
	\includegraphics[width=0.7\linewidth]{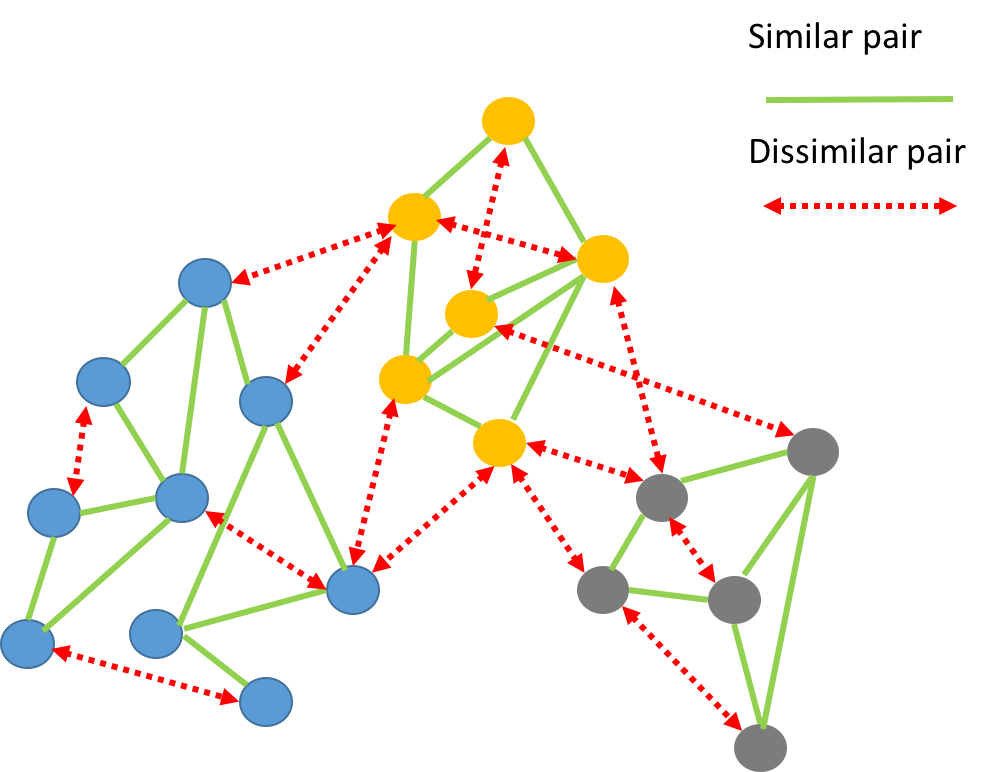}
	\caption{An example showing how categories can be discovered with dense pairwise constraints. Even though the dense pairwise constraints are predicted on a new domain and are therefore noisy, the genuine structure in the data is still able to be discovered.}
	\label{fig:clustering_example}
\end{figure}

Such a mapping of input data to a category distribution is difficult to learn in an unsupervised manner, however. Clustering techniques either use a strongly biased metric (e.g. Euclidean or cosine) or learn metrics by making various assumptions about the distributions in data. In our method, we posit that a correct mapping can be learned if a dense set of weak constraints (specifically, pairwise binary labels) are given. Further, these constraints can be learned in one domain and transferred to another domain. Despite the inherent noise in the predicted constraints, we posit that it will be possible to find structure in the data. In figure \ref{fig:clustering_example}, we illustrated an example where some pairwise constraints can be incorrectly predicted, but given a dense set of pairwise relationships we are still able to find the cluster structure from raw data. 

As described below, our specific instantiation of this idea learns a similarity metric using data from one domain, and transfers this metric to another domain. We then use this metric to generate pairwise constraints on unlabeled data in a new target domain, and use an robust deep-learning based clustering algorithm to estimate the categories in the data. Despite noise in these predicted constraints, we show that our clustering algorithm is able to accurately categorize the unlabeled data instances and achieve state of art results with respect to clustering metrics.

\subsection{Overall Framework Details}

\begin{figure*}	
	\begin{center}
		\includegraphics[width=0.8\textwidth]{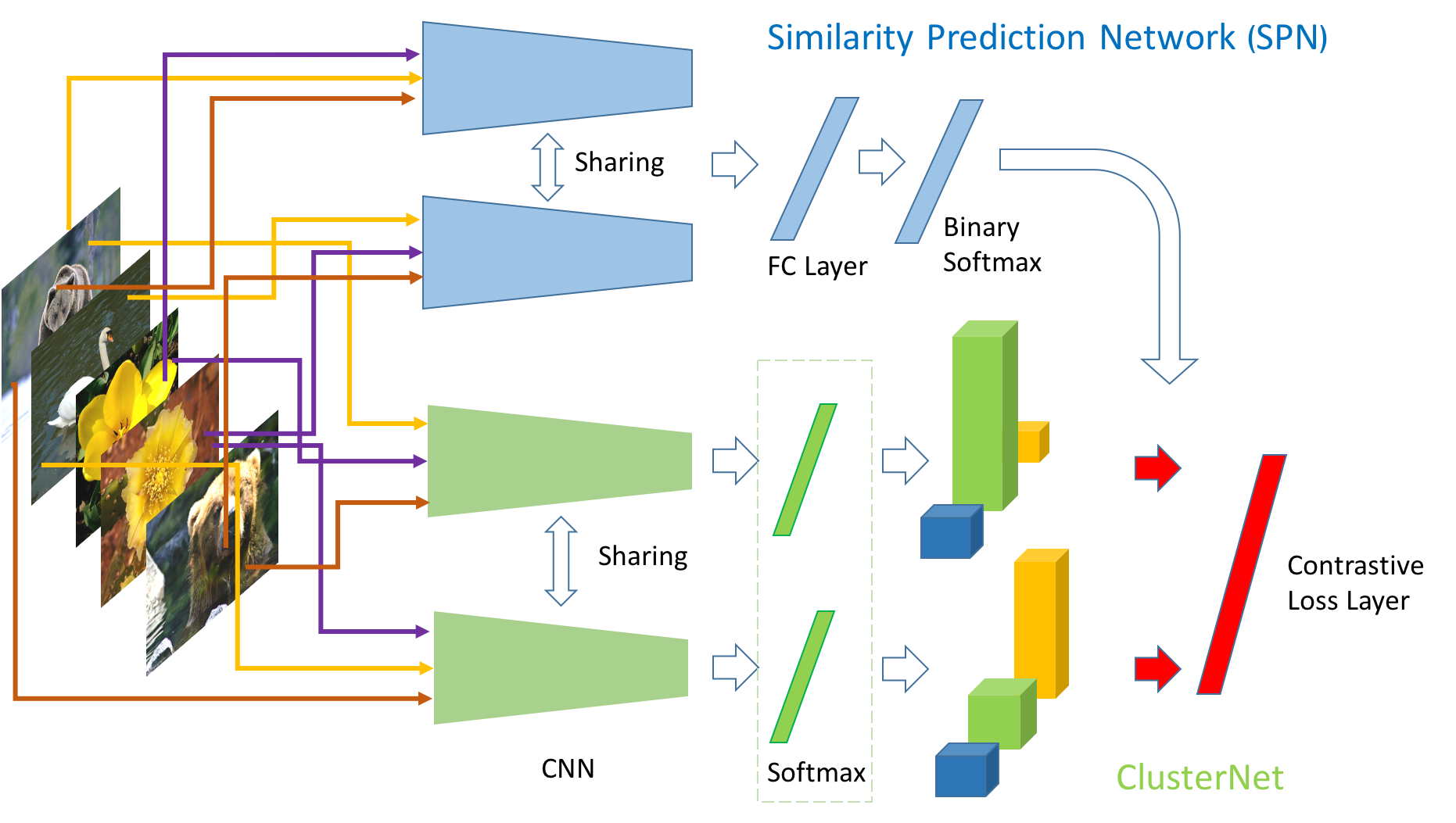}
	\end{center}
	\caption{Our overall network architecture. We densely sample pairs from the unlabeled images. For each pair $\image{p}, \image{q}$, the SPN will predict whether they are similar or dissimilar objects. The clustering network (ClusterNet) will predict the distribution over categories $\out{p}=\mathcal{P}, \out{q}=\mathcal{Q}$ or both pairs individually, and forward to the contrastive loss layer. During unsupervised clustering training, a contrastive loss layer will be used to backward-propagate gradients only to the ClusterNet (SPN is fixed) which learns the mapping from unlabeled images to a parametric distribution.}
	\label{fig:network}	
\end{figure*}


\subsubsection*{Discovering Categories using Contrastive Loss}

Similar to \cite{Hsu16iclrw}, we learn a category mapping through weak pair-wise constraints. Given a pair of images $\image{p}, \image{q}$, their corresponding output distributions are defined as $\out{p}=\mathcal{P}$ and $\out{q}=\mathcal{Q}$, which is described as a distribution over output categories. 

If the pair $\image{p}, \image{q}$ is a similar pair, they are expected to be mapped to the same category. The cost of a similar pair is described as the KL-divergence between the distributions:

\begin{align}
\loss{\image{p}, \image{q}}^{+} &= \KL{\mathcal{P}^{\star} || \mathcal{Q}} + \KL{\mathcal{Q}^{\star} || \mathcal{P}} \\
\KL{\mathcal{P}^{\star}||\mathcal{Q}} &= \sum_{\clusters} {p(\cluster{i}) log(\frac{p(\cluster{i})}{q(\cluster{i})})} \\
\mathcal{P}&=\neuralnetwork{\image{p}}, \mathcal{Q}=\neuralnetwork{\image{q}}
\end{align}

The cost $\loss{\image{p}, \image{q}}^{+}$ is symmetric w.r.t. $\image{p}, \image{q}$, in which $\mathcal{P}^{\star}$ and $\mathcal{Q}^{\star}$ are alternatively assumed to be constant.  Each KL-divergence factor $\KL{\mathcal{P}^{\star} || \mathcal{Q}}$ becomes a unary function whose gradient is simply $\partial \KL{\mathcal{P}^{\star} || \mathcal{Q}} / \partial \mathcal{Q}$. 

If $\image{p}, \image{q}$ comes from a pair which is defined as dis-similar, their output distributions are expected to be different, which can be defined as a hinge-loss function as: 

\begin{align}
\loss{\image{p}, \image{q}}^{-} &= \hingeloss{\KL{\mathcal{P}^{\star}||\mathcal{Q}}, \sigma}  + \hingeloss{\KL{\mathcal{Q}^{\star}||\mathcal{P}}, \sigma} \\
\hingeloss{e, \sigma} &= max(0, \sigma - e)
\end{align}

Given a pair with similarity constraint $l=\{0, 1\}$, the total loss can be defined as a contrastive loss as:

\begin{align}
\loss{\image{p}, \image{q}} = [l=1] \loss{\image{p}, \image{q}}^{+} + [l=0]\loss{\image{p}, \image{q}}^{-} 
\end{align}

in which $[\cdot]$ is an Iverson Bracket. Given a fixed constraint $l$ in the loss function, the contrastive loss can be end-to-end optimized with the clustering network.


\subsubsection*{Category Similarity Prediction Network}

The pairwise constraint per instance, $l$, is not available in unlabeled images. However, they can be predicted with a pre-trained network trained on another domain. We propose to use a similar weight-shared Siamese architecture as \cite{Zagoruyko15cvpr} to predict whether the two input images come from the same category.
	
The SPN is trained with an existing dataset from another domain that has category labels. During training, the input to this network is pairs of images, and the training objective is defined by whether they come from the same object category. Such a trained network is expected to have the ability to determine whether two given images are from a similar category or different categories. This simple form of output is used as the only transferred prior information, and can be used to train the clustering network using unlabeled data from the target domain. 
		
Figure \ref{fig:network} shows the entire training process of unsupervised object category. Given all of the unlabeled images, we densely sample pairs from them and predict whether each pair of images $\image{p}, \image{q}$ is a similar or dissimilar pair, thereby obtaining the label $l$. For the pair $\image{p}, \image{q}$, their output from ClusterNet are $\out{p}, \out{q}$. With the predicted label from SPN, they compose as a $(\out{p}, \out{q}, l_{p,q})$, feed as input to Contrastive Loss Layer. Such inputs can be easily fed into the CluserNet as mini-batches, which can be used to perform training efficiently. During training, errors are propagated from Contrastive Loss Layer to the ClusterNet, while SPN is fixed. During testing time, the ClusterNet predicts the distribution $\out{i}$ given a single image $\image{i}$. Through non-maximal suppression we can predict its corresponding category $\cluster{i}$.

\subsection{Efficiently Sampling Dense Pairwise Constraints within Mini-batches}


It is infeasible to predict similar or dissimilar pairs within a large dataset and train them with all possible pairs: the number of pairs will grow as a complexity of $\bigO{N^2}$, and training with multiple epochs will be very slow.

Instead, we choose to predict the dense pairwise similarity for samples within every mini-batch. In each epoch, the input images are randomly shuffled, and therefore generating mini-batches with random indices. Given images in each mini-batch, the SPN will estimate whether the label of each pair. Then all of the pairs are trained as processed in Figure \ref{fig:network}. This allows for extremely efficient large-scale clustering. 


\section{Diagnosing the Clustering Network}

Our framework contains two major components. The first one is a Siamese network that minimizes the contrastive loss for clustering. The second one is another Siamese network used to predict the binary similarity that is used in the contrastive loss. To understand what performance is required from the Similarity Prediction Network (SPN) for the clustering network to perform well, we simulate different similarity prediction performances by using ground truth labels. The precision and recall of the similar and dissimilar pairs were controlled to explore the clustering performance under different density and number of clusters.

\subsection{Evaluation Metric}
\label{sec:eval_metric}

To evaluate the performance of clustering, two standard metrics were used here and in the following sections. The first is \emph{unsupervised clustering accuracy} (ACC) \cite{yang2010image} and the second is \emph{normalized-mutual information} (NMI) \cite{strehl2002cluster}.

\subsubsection*{Unsupervised Clustering Accuracy (ACC):}

Each ground truth label is only assigned to one cluster in this measurement. If the number of clusters is larger than real categories number, there will be some clusters that are never assigned. Any samples that fall into the unassigned cluster will be regarded as an error. Given cluster $\Theta$ and the ground truth categories $\mathcal{C}$,  ACC measures the average accuracy as:

\begin{align} \label{eq:acc}
\textbf{ACC}(\Theta, \mathcal{C}) = \max_{f} \frac{\sum_{i} \mathbf{1} \{\theta_i = f(c_i)\}}{ N }  
\end{align}

where $N$ is the total number image number. $\max_{f}(\cdot)$ is the Hungarian algorithm which finds the best mapping between $\Theta$ and $\mathcal{C}$.

\subsubsection*{Normalized Mutual Information (NMI):}

\begin{align}
\textbf{NMI}(\Theta, \mathcal{C}) &= \frac{I(\Theta;\mathcal{C})}{\sqrt{H(\Theta)H(\mathcal{C})}} \\
I(\Theta;\mathcal{C}) &= \sum_{k}\sum_{j} \frac{ | \theta_{k} \cap c_{j} | }{N} \log(\frac{N | \theta_{k} \cap c_{j} |}{| \theta_{k} | | c_{j} |}) \\
H(\Theta) &= -\sum_k \frac{ | \theta_{k}| }{N} \log(\frac{ | \theta_{k}| }{N})
\end{align}

where $H(\Theta)$ represents entropy for $\Theta$ and $I(\Theta;\mathcal{C})$ represents mutual information. $N$ is the total number of images. 

\subsection{Experimental Setting}

To quickly explore the large combination of factors that may affect the clustering, we use a small dataset (MNIST) and a small network which has two convolution layers followed by two fully connected layers. The MNIST dataset is a dataset of handwritten digits that contains 60k training and 10k testing images with size 28x28. Only the training set is used in this section and the raw pixels, which were normalized to zero mean and unit standard deviation, were fed into networks directly.

The networks was randomly initialized and the clustering training was run five times under each combination of factors with showing the best final results, as is usual in the random restart regime. The mini-batch size was set to 256, thus up to 65536 pairs were presented to the contrastive loss per mini-batch if using full density (D=1). There were 235 mini-batches in an epoch and the optimization proceeded for 15 epochs. The clustering loss was minimized by stochastic gradient descent with learning rate 0.1 and momentum 0.9. The predicted cluster was assigned at the end by forwarding samples through the clustering networks. The best result in the five runs was reported.

To simulate different performance of the similarity prediction, the label of pairs were flipped according to the designated recall. For example, to simulate a 90\% recall of similar pair, 10\% of the ground truth similar pair in a mini-batch were flipped. The precision of similar/dissimilar pairs is a function of the recall of both type of pairs, thus controlling the recall is sufficient for the evaluation. The recalls for both similar and dissimilar pairs were gradually reduced from one to zero at intervals of 0.1.

\begin{figure}[t] 
	\centering
	\includegraphics[width=1.1\linewidth]{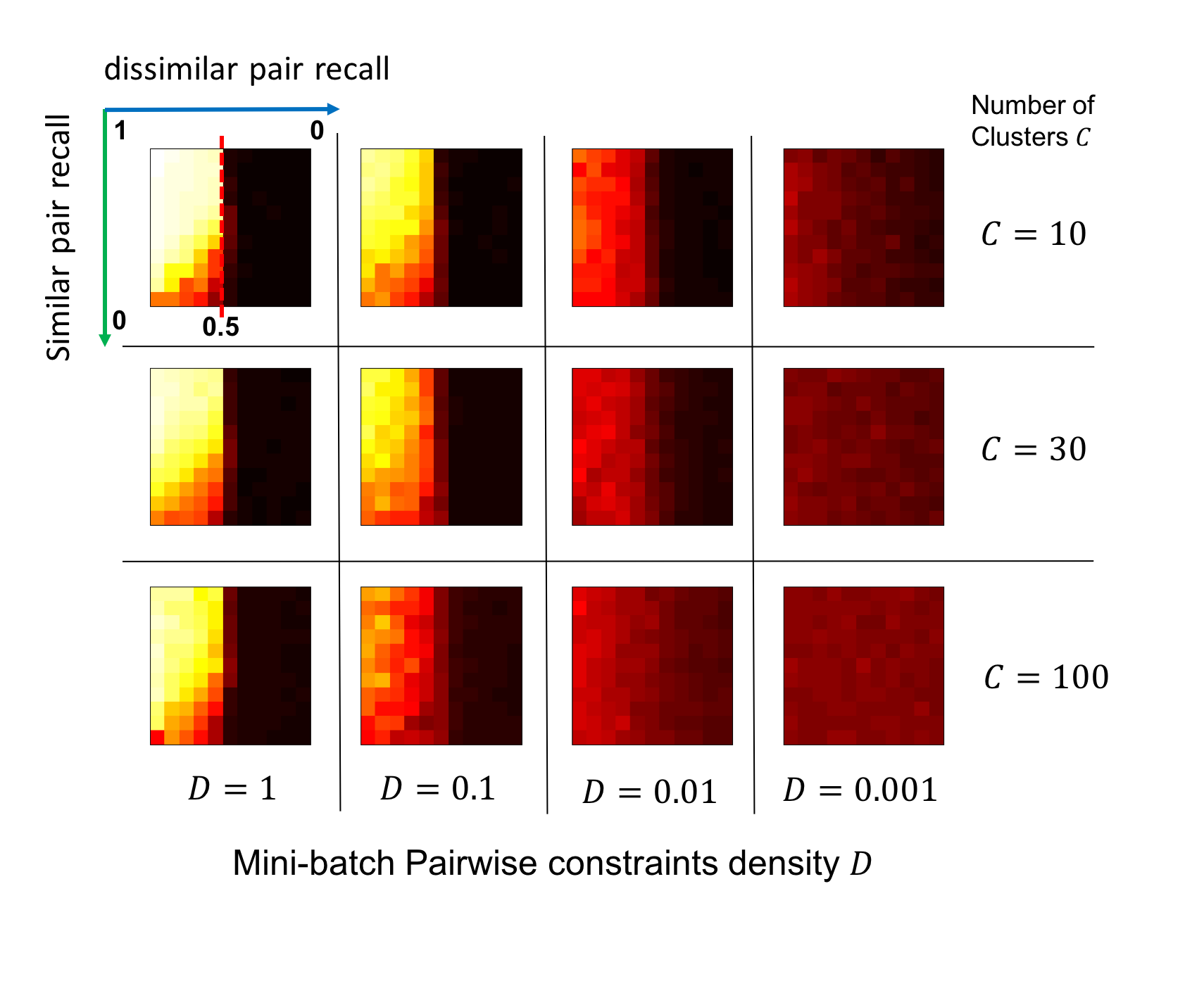}
	\caption{Clustering performance with different pairwise density and number of clusters. A bright color means that the NMI score is close to 1 while black corresponds to 0. The density is defined as a ratio compared to the total number of pair-wise combinations in a mini-batch. The number of clusters defines the final softmax output dimensionality. In each sub-figure, we show how the scores change w.r.t. the similar pair recall and dissimiliar pair recall.}
	\label{fig:parameter_discussion}
\end{figure}

\subsection{Discussion} 
\label{sec:concept_discussion}

The resulting performance w.r.t different values of recall, density, and number of clusters is visualized in Figure \ref{fig:parameter_discussion}. The bright color means high NMI score and is desired. The larger the bright region, the more robust the clustering is against the noise of similarity prediction. The ACC score shows almost the same trend and is thus not shown here.

\subsubsection*{How does similarity prediction affect clustering}

Looking at the top-left heat map in figure \ref{fig:parameter_discussion}, which has $D=1$ and 10 cluster, it can be observed that the NMI score is very robust to low similar pair recall, even lower than 0.5. 
For recall of dissimilar pairs, the effect of recall is divided at the 0.5 value: the clustering performance can be very robust to noise in dissimilar pairs if the recall is greater than 0.5; however, it can completely fail if recall is below 0.5. For similar pairs, the clustering works on a wide range of recalls when the recall of dissimilar pairs is high.

In practical terms, robustness to the recall of similar pairs is desirable because it is much easier to predict dissimilar pairs than similar pairs in real scenarios. In a dataset with 10 categories e.g. Cifar-10, we can easily get 90\% recall for dissimilar pairs with purely random guess, while the recall for similar pairs will be 10\%.

\subsubsection*{How does the density of the constraints affect clustering}

We argue that the density of pairwise relationships is the key factor to improving the robustness of clustering. The density $D=1$ means that every pair in a mini-batch is utilized by the clustering loss. For density $D=0.1$, it means only 1 out of 10 possible constraints is used. We could regard the higher density as better utilization of the pairwise information in a mini-batch, thus more learning instances contribute to the gradients at once. Consider a scenario where there is one sample associated with 5 true similar pairs and 3 false similar pairs. In such a case, the gradients introduced by the false similar pairs have a higher chance to be overridden by true similar pairs within the mini-batch, thus the loss can converge faster and is less affected by errors. In Figure \ref{fig:parameter_discussion}, we could see when density decreases, the size of the bright region shrinks significantly.

In our implementation, enumerating the full pairwise relationships introduces negligible overhead in computation time using GPU. Although there is overhead for memory consumption, it is limited because only the vector of predicted distributions has to be enumerated for calculating the clustering loss.

\subsubsection*{The effect of vary number of Cluster}

In the MNIST experiments, the number of categories is 10. We augment the softmax output number up to 100. The rows of figure \ref{fig:parameter_discussion} show that even when the number of output categories is significant larger than the number of true object categories, e.g. $100 > 10$, the clustering performance NMI score only degrades slightly. 


\section{Transfering similarity to unlabeled datasets} \label{sec:experiments}


This section presents the experimental results of learning the similarity on one dataset and then predicting on another dataset for contrastive loss clustering. To show robustness to prediction error and demonstrate positive transfer, both the performance of SPN and ClusterNet will be shown. 

\subsection{Transfer From Omniglot to Mnist}
Omniglot \cite{graves2016hybrid} is a handwritten dataset containing 50 alphabets with a total of 1623 different characters. The dataset was split to 30 and 20 alphabets for background set and evaluation set, which has 964 and 659 characters, respectively. Examining the transferability of the learned model to Mnist is a common practice in some one-shot learning tests \cite{Vinyals16nips,Koch15icmlw}.

We use the Omniglot background set for training the SPN. The images were resized to 32x32, normalized to zero mean and unit standard deviation. Random resized cropping was used for data augmentation. The SPN has four basic blocks consisting of a 3x3 convolution layer, a batch normalization layer, and a rectifier linear unit, followed by a 2x2 max pooling layer. There are two fully connected layers after the four convolution layers and the last hidden layer has the output dimension 512. Another two fully connected layers were added on top of a pair of the basic networks. The training criterion used cross entropy for two class classification (similar and dissimilar). We also utilized the full pair-wise relationships in a mini-batch for training. The Siamese architecture of SPN was used during both training and prediction.

The ClusterNet has the same basic structure as SPN. The difference is that a hinged KL-divergence loss is used for the criterion. The Siamese architecture is only used during optimization of the clustering loss. After that, only the basic network which has four convolution and two fully connected layers is needed for predicting the cluster by feed-forwarding a sample through ClusterNet. The implementation is on Torch and the code is available upon publication. 

When using a threshold of 0.5 to binarize the two-class prediction, the trained SPN performance was 0.659 recall for similar pairs, and 0.892 recall for dissimilar pairs. While the recall is not high, when we use the N-way test which is commonly used to evaluate the performance of one-shot learning, our SPN outperform other similarity learning methods when transferring to the MNIST test set with a significant margin (see Table \ref{tab:N-way_similarity}). This shows that training with dense pairs improves the performance significantly.

To show that positive transfer happened while transferring the similarity prediction, we compare our clustering algorithm with different state-of-the-art unsupervised clustering approaches: K-means \cite{Macqueen67bsmsp}, N-Cuts \cite{Shi97pami}, SC-LS \cite{chen2011large}, NMF-LP  \cite{Cai09ijcai}, JULE\cite{Yang16cvpr} and DEC\cite{Xie16icml}. The implementation and best results are based on both \cite{Yang16cvpr} and \cite{Xie16icml}. Table \ref{tab:mnist_clustering} shows our approach has a significant advantage. The ACC is even able to reach the performance of a supervised classification task.

\begin{table} 
	\centering
	\caption{N-way test on Omniglot\cite{Lake15science} and MNIST, with accuracy comparison with state-of-art similarity learning methods.}
	\label{tab:N-way_similarity}
	\begin{tabular}{@{}lccc@{}}
		\toprule
		& \multicolumn{2}{c}{Omniglot-eval}    & \multicolumn{1}{l}{MNIST-test} \\
		N-way Accuracy        & 5-way          & 20-way         & 10-way                         \\  \midrule
		Siamese-Nets \cite{Koch15icmlw} & 0.967    & 0.880   & 0.703                          \\
		Match-Net \cite{Vinyals16nips}  & \textbf{0.981} & \textbf{0.938} & 0.720    \\ \midrule
		Ours (Omniglot-bg)    & 0.979          & 0.935          & \textbf{0.794}                 \\ \bottomrule
	\end{tabular}
\end{table}

\begin{table}[h]
	\centering
	\caption{Category Clustering Performance over MNIST with State-of-art clustering approaches.$*$:The similarity is trained on Omniglot. $**$:The similarity is trained on Cifar100. \textasciicircum:The data is simply fed-forward into the ClusterNet which was optimized with train-set. }
	\label{tab:mnist_clustering}
	\begin{tabular}{@{}lccc@{}}
		\toprule
		\multicolumn{1}{c}{NMI} & \multicolumn{1}{l}{MNIST-train} & \multicolumn{1}{l}{MNIST-test} & \multicolumn{1}{l}{Cifar10-test} \\ \midrule
		K-means\cite{Macqueen67bsmsp} & 0.500       & 0.528      & 0.085                 \\
		N-Cuts \cite{Shi97pami}                 & 0.411       & 0.386   & 0.040                    \\
		SC-LS \cite{chen2011large}                & 0.706      & 0.756   & 0.096                    \\
		NMF-LP  \cite{Cai09ijcai}                & 0.452      & 0.467  & 0.071                     \\
		JULE\cite{Yang16cvpr} & 0.913                           & 0.915   & 0.130                  \\ \midrule
		\textbf{Ours*} 		& \textbf{0.966}              & \textbf{0.974\^}  & -       \\ 
		\textbf{Ours**} 		& -              & -  & \textbf{0.403\^}       \\ 
		\midrule
		\multicolumn{1}{c}{ACC} & \multicolumn{1}{l}{MNIST-train} & \multicolumn{1}{l}{MNIST-test} & \multicolumn{1}{l}{Cifar10-test} \\
		\midrule
		DEC\cite{Xie16icml} 	& 0.843						     & -	& -						\\ \midrule
		\textbf{Ours*} 		& \textbf{0.988}              & \textbf{0.990\^}  & -      \\
		\textbf{Ours**} 		& -              & -  & \textbf{0.435\^} \\ \bottomrule
	\end{tabular}
\end{table}

\subsection{From Cifar100 to Cifar10}

The Cifar100 and Cifar10 \cite{krizhevsky09} datasets both contain color images of size 32x32 and have 100 and 10 categories, respectively. There is no overlap between the two datasets in the fine level of the cifar100 category, and thus can be used as a good target to validate transfer learning. The setup of networks is the same as the Omniglot-Mnist experiment. Table \ref{tab:mnist_clustering} shows that clustering on Cifar10 is very difficult. Even the JULE approach which could jointly learn the feature space does not work well on the dataset. The result also reveals the difficulty of using clustering to discover object categories from raw pixel.

\subsection{Image Category Discovery over Natural Images}

\begin{figure}
\centering
\includegraphics[width=0.7\linewidth]{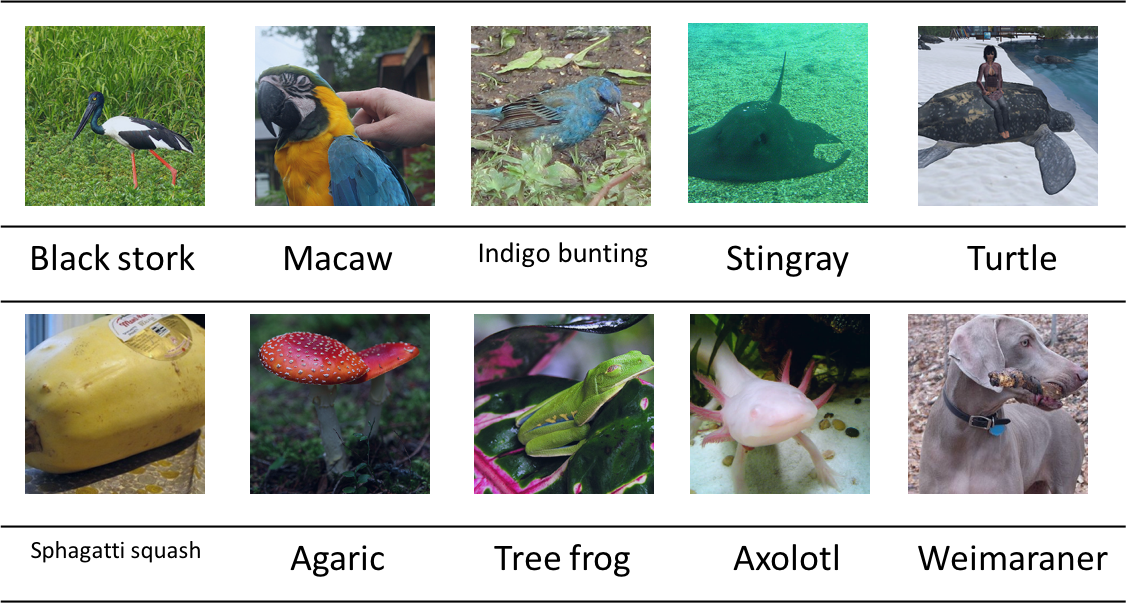}
\caption{The 10 discovery categories as imagenet-10R.}
\label{fig:imagenet_examples}
\end{figure}

To demonstrate our approach on natural images, we randomly exclude 118 categories from ImageNet as \cite{Vinyals16nips} to train the SPN. We expect to discover 10 categories, randomly selected from the hold out categories, which is shown in Figure \ref{fig:imagenet_examples}. Some closely related categories like Black stork and Indigo bunting are included thus increasing the difficulty of differentiating them into different clusters.

The SPN used in this section has Resnet-50 \cite{He2015} as the basic network and its weights were randomly initialized. For showing that positive transfer happened, we use the features from the ImageNet-pretrained VGG-16 \cite{simonyan2014very} networks for the clustering method. While the pre-trained features may be affected by instances in Imagenet-10R, not every clustering algorithm is able to use this feature space to separate the clusters into different categories. Thus it is meaningful to do the comparison using the pretrained features. Additional conditions varying the sources used for training the SPN and initializing the ClusterNet can be found in the supplementary material.

Our approach still shows strong advantage over other clustering algorithms in Table \ref{tab:imagenet_clustering_vs_pretrained_features}. The visualization of the clustering results was shown in Figure \ref{fig:imagenet_visualization}, which used the predicted cluster distribution as the feature vector for t-SNE. Most images of the same category are in close vicinity. Even the SPN-unseen categories such as Black stork and Indigo bunting are separated well, thus proves the validity of our approach.

The STL-10 dataset \cite{coates2011analysis} is a subset of ImageNet with image size 96x96. However it only contains higher level categories, thus could be used as a counter dataset to Imagenet-10R which has some fine-grained classes. Our approach is also shown to achieve top performance on it in Table \ref{tab:imagenet_clustering_vs_pretrained_features}.


\begin{table} [t]
	\centering
	\caption{Clustering evaluation on natural images. }
	\label{tab:imagenet_clustering_vs_pretrained_features}
	\begin{tabular}{@{}clcccc@{}}
		\toprule
		\multicolumn{1}{l}{}                                                          &         & \multicolumn{2}{c}{Train-set}                                                & \multicolumn{2}{c}{Test-set}                                                 \\
		\multicolumn{1}{l}{}                                                          &         & NMI                                & ACC                                 & NMI                                & ACC                                 \\  \midrule
		\multirow{5}{*}{\begin{tabular}[c]{@{}c@{}}Imagenet\\10R\end{tabular}} & K-means\cite{Macqueen67bsmsp} & 0.600                              & 0.586                             & 0.426                              & 0.376                              \\
		& N-Cuts \cite{Shi97pami}    & 0.127                              & 0.172                              & 0.131                              & 0.170                              \\
		& NMF-LP  \cite{Cai09ijcai}  & 0.515                              & 0.603                              & 0.530                              & 0.552                              \\
		& SC-LS \cite{chen2011large}   & 0.604                              & 0.648                              & 0.267                              & 0.332                              \\
		& Ours    & \textbf{0.691}                     & \textbf{0.759}                     & \textbf{0.707\^}                     & \textbf{0.750\^}                     \\ \midrule
		\multirow{5}{*}{STL-10}                                                       & K-means\cite{Macqueen67bsmsp} & \multicolumn{1}{l}{0.521}          & \multicolumn{1}{l}{0.538}          & \multicolumn{1}{l}{0.512}          & \multicolumn{1}{l}{0.458}          \\
		& N-Cuts \cite{Shi97pami}   & \multicolumn{1}{l}{0.144}          & \multicolumn{1}{l}{0.173}          & \multicolumn{1}{l}{0.149}          & \multicolumn{1}{l}{0.177}          \\
		& NMF-LP  \cite{Cai09ijcai}  & \multicolumn{1}{l}{0.454}          & \multicolumn{1}{l}{0.481}          & \multicolumn{1}{l}{0.432}          & \multicolumn{1}{l}{0.428}          \\
		& SC-LS \cite{chen2011large}   & \multicolumn{1}{l}{0.518}          & \multicolumn{1}{l}{0.515}          & \multicolumn{1}{l}{0.509}          & \multicolumn{1}{l}{0.525}          \\
		& Ours    & \multicolumn{1}{l}{\textbf{0.649}} & \multicolumn{1}{l}{\textbf{0.721}} & \multicolumn{1}{l}{\textbf{0.639\^}} & \multicolumn{1}{l}{\textbf{0.711\^}} \\ \bottomrule
	\end{tabular}
\end{table}

\section{Conclusion}
We present a novel approach to performing object category discovery by transfering image similarity prediction from a source domain to facilitate unsupervised clustering on a different target domain where categories are unknown and data is unlabeled. The experiments show that this approach is not only valid on simple image datasets such as Omniglot and MNIST, but also scales to and works on the ImageNet dataset. The proposed clustering method is robust to noisy similarity predictions due to the densely-sampled similarities used for clustering. In future work, we will extend the clustering method to datasets with a larger number of categories.

\section*{Acknowledgments}
This work was supported by the National Science Foundation and National Robotics Initiative (grant \# IIS-1426998).

{\small
\bibliographystyle{ieee}
\bibliography{refMacro,egbib}
}

\newpage
\appendix
\section{Appendix: Varying the Source of SPN and ClusterNet}
To demonstrate that our approach works across different random initializations of the experiment, we randomly picked another 882 categories from ImageNet to train the SPN, and randomly used 10 (ImageNet-10R2) out from the hold-out 118 categories for clustering. The ClusterNet consists of the basic network, specifically the CNN and fully connected layers, and was pretrained under several conditions. The effects of using different ways to initialize the ClusterNet are summarized in Table \ref{tab:imagenet_clustering_run2}. Due to the limited number of images in the target set for clustering, optimizing the clustering loss with very large networks such as VGG16 and Resnet were not feasible. Thus a better initialization for ClusterNet was necessary. We therefore used an unsupervised feature learning approach introduced by \cite{doersch2015unsupervised}, which learned an image patch representation by predicting the relative location of patches. In this way the supervision involved in clustering was minimized.

\begin{table*}
	\centering
	\caption{ImageNet-10R2 clustering results. \emph{VGG-1000}: The features are trained from 1000-classes classification. \emph{SPN-882}: From the SPN network trained with 882 classes. \textasciicircum:The data is simply fed-forward into the ClusterNet which was optimized with train-set. \emph{Finetune CNN}: The convolution layers are finetuned or not.}
	\label{tab:imagenet_clustering_run2}
	\begin{tabular}{llcllll}
		\multirow{2}{*}{\begin{tabular}[c]{@{}l@{}}Feature/\\ClusterNet initialization\end{tabular}}        & \multirow{2}{*}{Method} & \multirow{2}{*}{\begin{tabular}[c]{@{}c@{}}Finetune\\CNN\end{tabular}} & \multicolumn{2}{c}{Train-set} & \multicolumn{2}{c}{Test-set} \\
		&           &              & NMI           & ACC           & NMI           & ACC          
		\\ \midrule
		\multirow{5}{*}{\begin{tabular}[c]{@{}l@{}}Unsupervised feature\\ (Context\cite{doersch2015unsupervised})\end{tabular}} & K-means\cite{Macqueen67bsmsp}                & - & 0.288              & 0.384              & 0.299              & 0.382             \\
		& NMF-LP\cite{Cai09ijcai}                 & - & 0.328              & 0.264              & 0.340              & 0.438             \\
		& SC-LS\cite{chen2011large}                 & - & 0.335              & 0.429              & 0.310              & 0.386             \\
		& Ours                  & No & 0.516              & 0.619              & 0.519\textasciicircum              & 0.602\textasciicircum             \\
		& Ours                  & Yes & \textbf{0.674}              & \textbf{0.718}              & \textbf{0.661\textasciicircum}              & \textbf{0.700\textasciicircum}             \\
		\midrule
		\begin{tabular}[c]{@{}l@{}}Supervised feature\\ (VGG-1000)\end{tabular}               & Ours                  & No & 0.770              & 0.837              & 0.761\textasciicircum              & 0.824\textasciicircum             \\
		\midrule
		\begin{tabular}[c]{@{}l@{}}Supervised feature\\ (SPN-882)\end{tabular}                        & Ours                  & Yes & 0.653              & 0.658              & 0.653\textasciicircum              & 0.670\textasciicircum             \\
		\midrule
	\end{tabular}
\end{table*}

\subsection{Discussions}
The performance of using supervised features in Table \ref{tab:imagenet_clustering_run2} could be regarded as an approximated upper bound for judging the clustering performance. By using an unsupervised feature learning approach to initialize the parameters of the ClusterNet, our approach outperformed other state-of-the-art clustering methods by $\sim60\%$ in terms of NMI without fine-tuning the convolution layers, and $\sim100\%$ with finetuning, which only has a difference of $\sim15\%$ from using supervised features. In the unsupervised feature learning case (with or without fine-tuning), note that we do not transfer any supervised features from the source domain (i.e. we only use the similarity network output) and use only unlabeled data in the target domain, yet we significantly outperform other clustering methods. Also note that when we perform fine-tuning, we update the convolution features of the ClusterNet but this is done purely using the similarity function outputs and no additional information (e.g. labels) are used for the target domain.

\begin{figure}	[b]
	\begin{center}
		\includegraphics[trim={5cm 5cm 5cm 5cm},clip,width=0.5\linewidth]{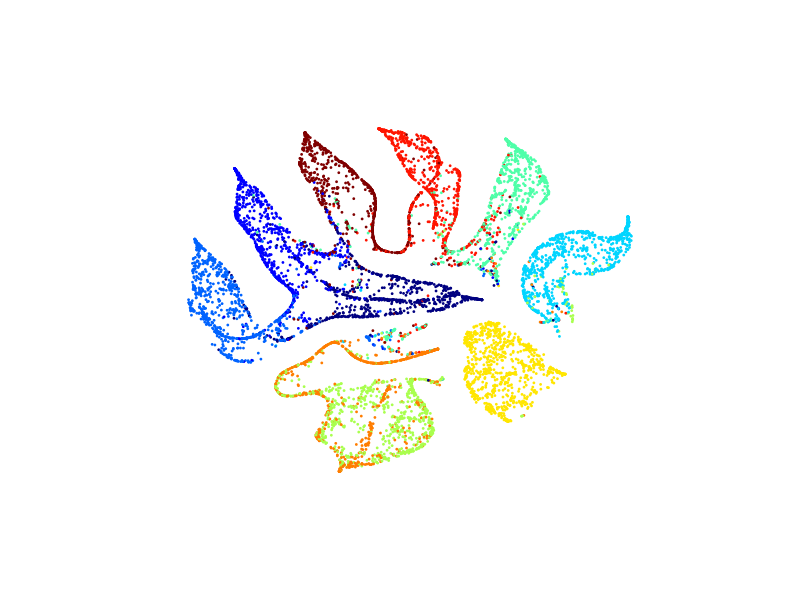}
	\end{center}
	\caption{The t-SNE visualization of clustering results using the supervised VGG-1000 feature in Table \ref{tab:imagenet_clustering_run2}. Each dot is colored by ground-truth label.}
	\label{fig:imagenet10_vis_label}	
\end{figure}

\begin{figure*}	
	\begin{center}
		\includegraphics[width=1\textwidth]{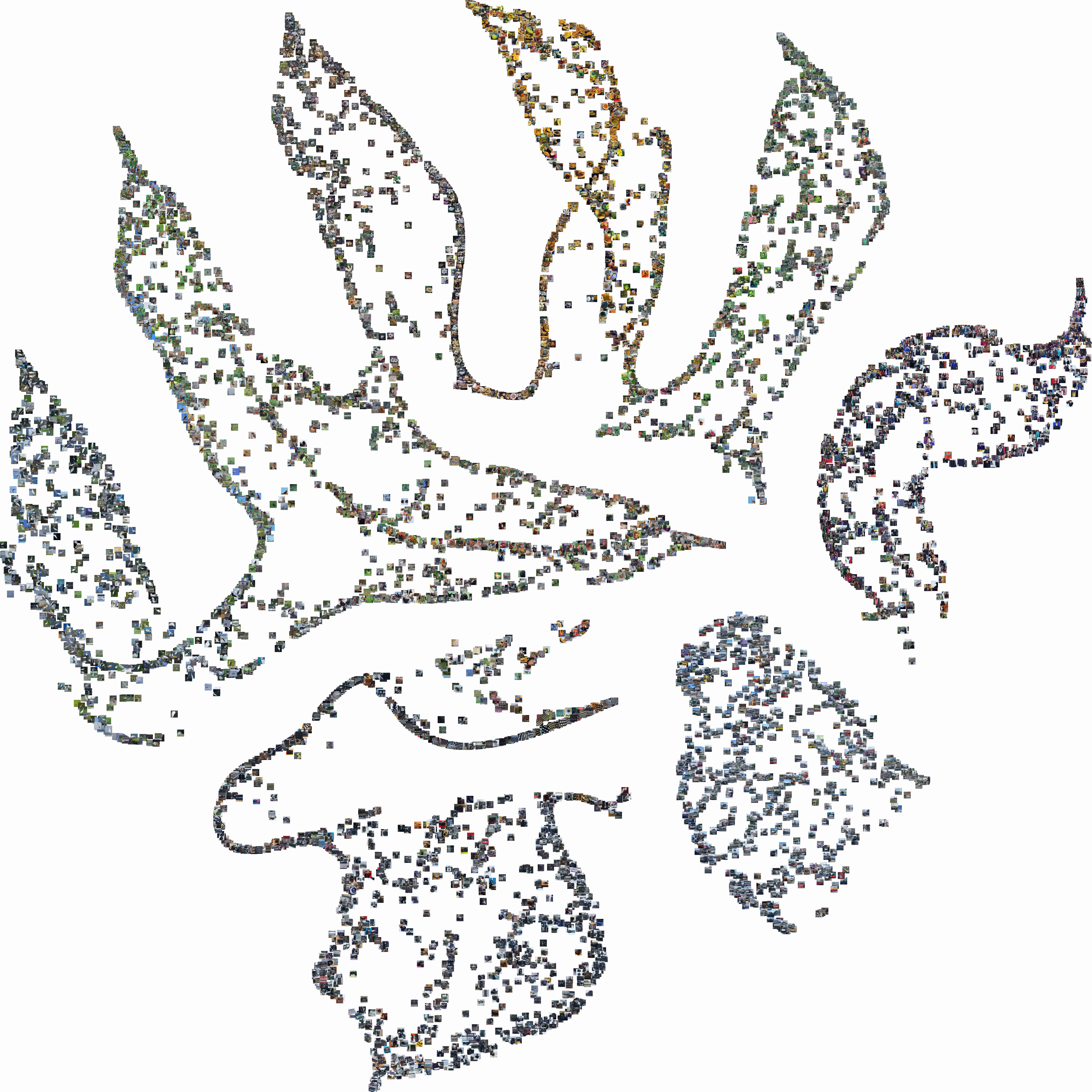}
	\end{center}
	\caption{The Visualization with images for Figure \ref{fig:imagenet10_vis_label}. The images are provided in fine resolution.}
	\label{fig:imagenet10_vis}	
\end{figure*}

The t-SNE visualization of the ImageNet-10R2 are shown in Figures \ref{fig:imagenet10_vis_label} and \ref{fig:imagenet10_vis}. The results are similar to Figure 1 in the paper, but the images are provided with fine resolution here for inspection. One thing worth noting is that the close categories, e.g. birds, are also close in vicinity, thus the predicted cluster distribution not only separates the categories but also maintains the distance between categories. Another point to note is the striped structure in Figure \ref{fig:imagenet10_vis}. We argue that this pattern represents the trajectories transitioning from one category to another, such that they can be close to each other because of some common attributes (in terms of the outputs of last hidden layers). In contrast, for completely different categories such transitions do not exist in the visualization.

\subsection{Experimental Setting}
The experimental setup is very similar to the Section \emph{Image Category Discovery over Natural Images} in the paper. This section will provide more information to supplement the details that did not fit into the limited paper size. The difference in the supplementary experiments will also be pointed out.

\subsubsection{Training the SPN}
The basic network used here was Resnet-50, which has 49 convolution layers followed by an average pooling layer. The mini-batch contains 50 ImageNet images from 5 randomly sampled category in the 882-category training set. The output feature dimension of Resnet-50 is 2048. The feature vectors from the siamese Resnet-50 is concatenated into 4096 dimension in the first fully-connected layer of SPN and outputs a 8192D vector. Another fully-connected layer then converts it to 2D outputs and was followed by a softmax layer. Thus the output is the probability distribution across two classes: similar or dissimilar. A standard cross entropy criterion is used to calculate the loss. The full pairwise relationships in a mini-batch were enumerated before the feature concatenation, thus the batch size in the fully-connected layers became $50*50=2500$. For the preprocessing, all images were resized to 224x224 with standard normalization and augmented by random-sized cropping, color jittering, and horizontal flipping. The training proceeded with learning rate 0.1 and momentum 0.9 for 30 epochs, and then the learning rate was dropped to 0.01 for another 10 epochs.

\subsubsection{Training the ClusterNet}

The parameters of the ClusterNet can be initialized via various methods and be fine-tuned by optimizing the clustering loss. The initializations are described below.  

\subparagraph*{Unsupervised feature learning:}
We use the approach proposed by \cite{doersch2015unsupervised} and use the VGG-Style Network pretrained by the author. To construct the basic network, we only use the fully convolutional layers. Three fully-connected layers were sequentially added to convert the 7x7 feature map into 4096D, 4096D, and 10D outputs vector. A softmax layer was also added to convert the 10D outputs to a probability distribution. The size of a mini-batch is 50, thus there will be $50*50=2500$ pairs of distributions used in calculating the clustering loss. The images were purely randomly-selected from the training set of ImageNet-10R or ImageNet-10R2, which both contain 13000 images from 10 classes. The test-set is the corresponding subset in the ImageNet Validation set, which has 500 images in total. The image preprocessing is the same as SPN. To get the similarity prediction, the SPN ran at the same time using the same mini-batch in evaluation mode. The prediction is a probability between 0 and 1 and were binarized by a threshold of 0.5, where larger than 0.5 represented a similar pair. The resulting prediction was then used in the hinged KL-divergence criterion. The margin used in the clustering loss was fixed to 2 across all experiments, which is the value suggested in \cite{Hsu16iclrw}. The clustering proceeded for 80 epochs with a learning rate 0.01 if the CNN fine-tunning was disabled, and 0.001 if enabled. The learning rate was divided by 10 after 50th epoch. After the optimization stage, all train-set and test-set images were fed forward to ClusterNet again to obtain the final predicted cluster distribution.

\subparagraph*{VGG-1000:}
The VGG-1000 is the VGG16 network in \cite{simonyan2014very} trained with 1000-classes classification task on ImageNet. We use the networks in a similar way as described in last subsection. The outputs of the fifth convolution layer is also used as the feature for other unsupervised clustering algorithms. This initialization is also used for evaluating the capability of our approach and demonstrating the potential upper bound performance.

\subparagraph*{SPN-882:}
The SPN-882 is the networks trained in section \emph{Training SPN}. The basic network was extracted and copied to the basic network of ClusterNet. An extra linear layer was added to convert from the feature dimension to the 10 outputs. The training procedure is the same as the paragraph in \emph{Unsupervised feature}.

\subparagraph*{The list of ImageNet Random-118 used in Supplementary Experiments:}
 'n01443537'
 ,'n01484850'
 ,'n01514859'
 ,'n01530575'
 ,'n01558993'
 ,'n01560419'
 ,'n01614925'
 ,'n01694178'
 ,'n01728572'
 ,'n01737021'
 ,'n01751748'
 ,'n01843065'
 ,'n01943899'
 ,'n01978455'
 ,'n01980166'
 ,'n01983481'
 ,'n01984695'
 ,'n02009912'
 ,'n02058221'
 ,'n02086646'
 ,'n02087394'
 ,'n02089973'
 ,'n02091032'
 ,'n02092002'
 ,'n02093428'
 ,'n02093859'
 ,'n02095889'
 ,'n02096585'
 ,'n02098286'
 ,'n02099429'
 ,'n02102040'
 ,'n02105162'
 ,'n02106382'
 ,'n02107142'
 ,'n02108089'
 ,'n02108551'
 ,'n02108915'
 ,'n02112018'
 ,'n02113186'
 ,'n02113978'
 ,'n02127052'
 ,'n02128925'
 ,'n02129165'
 ,'n02165456'
 ,'n02231487'
 ,'n02396427'
 ,'n02412080'
 ,'n02423022'
 ,'n02483362'
 ,'n02486410'
 ,'n02487347'
 ,'n02492660'
 ,'n02493509'
 ,'n02669723'
 ,'n02730930'
 ,'n02793495'
 ,'n02795169'
 ,'n02804414'
 ,'n02807133'
 ,'n02870880'
 ,'n02916936'
 ,'n02951585'
 ,'n03100240'
 ,'n03124170'
 ,'n03131574'
 ,'n03187595'
 ,'n03188531'
 ,'n03216828'
 ,'n03255030'
 ,'n03272562'
 ,'n03379051'
 ,'n03400231'
 ,'n03447447'
 ,'n03457902'
 ,'n03496892'
 ,'n03534580'
 ,'n03594945'
 ,'n03658185'
 ,'n03662601'
 ,'n03666591'
 ,'n03691459'
 ,'n03697007'
 ,'n03782006'
 ,'n03786901'
 ,'n03794056'
 ,'n03857828'
 ,'n03874599'
 ,'n03944341'
 ,'n03991062'
 ,'n04026417'
 ,'n04033995'
 ,'n04065272'
 ,'n04069434'
 ,'n04070727'
 ,'n04131690'
 ,'n04162706'
 ,'n04201297'
 ,'n04311174'
 ,'n04325704'
 ,'n04344873'
 ,'n04371430'
 ,'n04372370'
 ,'n04380533'
 ,'n04487081'
 ,'n04501370'
 ,'n04505470'
 ,'n04522168'
 ,'n04540053'
 ,'n04579432'
 ,'n07579787'
 ,'n07717410'
 ,'n07734744'
 ,'n07760859'
 ,'n07802026'
 ,'n07873807'
 ,'n13037406'
 ,'n13044778'
 ,'n13052670'

\subparagraph*{The list of ImageNet Random-10 used in Supplementary Experiments:}
'n01514859'
,'n01558993'
,'n02009912'
,'n02669723'
,'n03991062'
,'n04069434'
,'n04487081'
,'n04505470'
,'n07717410'
,'n13044778'

\end{document}